\documentclass[]{llncs}
\usepackage[T1]{fontenc}
\usepackage{todonotes}
\usepackage{graphicx}
\usepackage[utf8]{inputenc} 
\usepackage{hyperref}    
\usepackage{url}            
\usepackage{booktabs}       
\usepackage{amsfonts}       
\usepackage{nicefrac}       
\usepackage{microtype}      
\usepackage{xcolor}         

\usepackage{amsmath}
\usepackage{cleveref}
\usepackage{algorithm}
\usepackage[noend]{algpseudocode}
\usepackage{subcaption}
\usepackage{tabularray}
\usepackage[figuresright]{rotating}
\usepackage{censor}

\usepackage{tikz}
\usetikzlibrary{calc}
\usetikzlibrary{positioning}
\usetikzlibrary{3d,decorations.text,shapes.arrows,fit,backgrounds,arrows.meta}
\usepackage{standalone} 
\usepackage{tikzscale} 
\usepackage{xspace}
\usepackage{mdframed}
\usepackage{multicol}

\usepackage{svg}
\usepackage{pgfplots}
\usepackage{graphicx}
\usepackage{subcaption}
\newtheorem{problemdef}{Problem}
\newenvironment{problembox}{\begin{mdframed}[backgroundcolor=black!10] \begin{problemdef}}{\end{problemdef}\end{mdframed}}

\newcommand{\sycam}[0]{\textsc{SyCAM}\xspace}

\begin{document}
\title{Metric-Guided Synthesis \\for Class Activation Mapping\thanks{This work was partly supported by the Wallenberg AI, Autonomous Systems and Software Program funded by the Knut and Alice Wallenberg Foundation, a Royal Academy of Engineering Research Fellowship and a Royal Society Industry Fellowship.}}

%
%
\author{Alejandro Luque-Cerpa\inst{1}\and
Elizabeth Polgreen  \inst{2} \and
Ajitha Rajan\inst{2} \and
\\Hazem Torfah\inst{1}}

%
\institute{Chalmers University of Technology and University of Gothenburg, Sweden \\ \email{\{luque, hazemto\}@chalmers.se} \and University of Edinburgh, UK \\ \email{elizabeth.polgreen@ed.ac.uk, arajan@ed.ac.uk}}
%

\maketitle              
\begin{abstract}
Class activation mapping (CAM) is a widely adopted class of saliency methods used to explain the behavior of convolutional neural networks (CNNs). These methods generate heatmaps that highlight the parts of the input most relevant to the CNN output. Various CAM methods have been proposed, each distinguished by the expressions used to derive heatmaps. In general, users look for heatmaps with specific properties that reflect different aspects of CNN functionality. These may include similarity to ground truth, robustness, equivariance, and more. Although existing CAM methods implicitly encode some of these properties in their expressions, they do not allow for variability in heatmap generation following the user's intent or domain knowledge.
In this paper, we address this limitation by introducing SyCAM, a metric-based approach for synthesizing CAM expressions. Given a predefined  evaluation metric for saliency maps, SyCAM automatically generates CAM expressions optimized for that metric. We specifically explore a syntax-guided synthesis instantiation of SyCAM, where CAM expressions are derived based on predefined syntactic constraints and the given metric.
Using several established evaluation metrics, we demonstrate the efficacy and flexibility of our approach in generating targeted heatmaps. We compare SyCAM with other well-known CAM methods on three prominent models: ResNet50, VGG16, and VGG19.

\end{abstract}

\section{Introduction}
Convolutional Neural Networks (CNNs) have enabled the development of efficient solutions for a wide range of challenging vision problems, such as object detection in autonomous vehicles \cite{6909475}, facial recognition through semantic segmentation \cite{554195}, and medical image analysis \cite{medicalcnn,rafferty2024transparentclinicallyinterpretableai}.
Despite their advantages, CNNs, like other neural models, suffer from an opaque decision-making process, making it challenging to build trust in their predictions.
For instance, physicians using a CNN for X-ray classification require more than just a diagnosis; they need to understand which specific part of the X-ray led to that diagnosis. Similarly, in autonomous vehicles, debugging and employing CNN-based object detectors requires insight into which parts of an image triggered a classification.
The growing need to explain CNN behavior has led to the development of various explainability techniques \cite{GradCAM++,ablationcam,scoreCAM,GradCAM}. However, with each new method, it is becoming increasingly clear that more systematic approaches are needed to generate explanations that adapt to the specific intents and needs of end users \cite{survey}.  

\begin{figure}[t]
    \centering
    \scalebox{0.8}{
    \includegraphics[width=\linewidth]{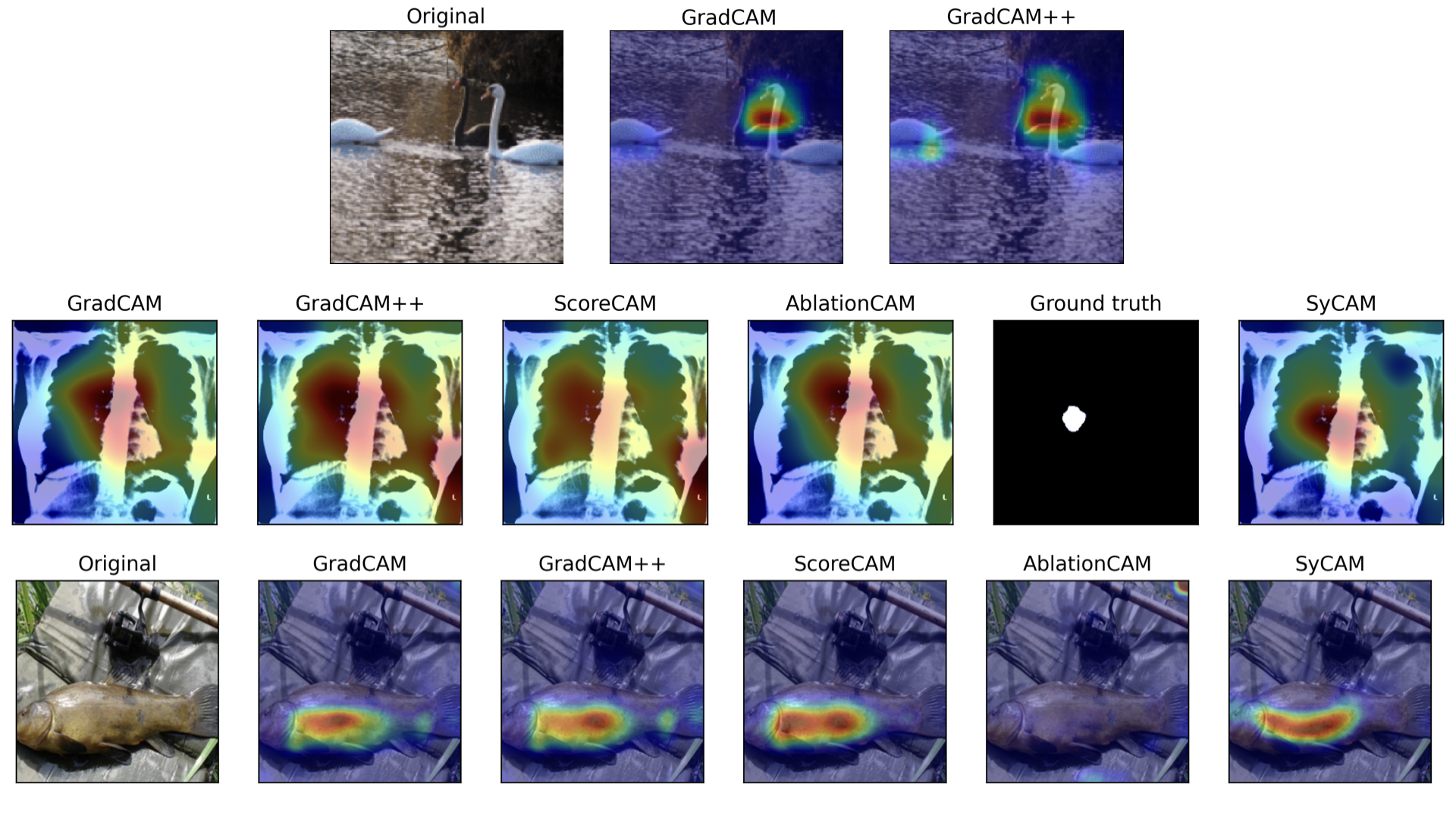}
    }
    \caption{\small 
    Saliency maps generated using different CAM methods (GradCAM \cite{GradCAM}, GradCAM++ \cite{GradCAM++}, ScoreCAM \cite{scoreCAM}, AblationCAM \cite{ablationcam}, and one using an expression synthesized by our  \sycam framework) for three models different CNNs trained on three different data sets ImageNet \cite{imagenet}, COVID-QU-Ex \cite{covid_dataset} and ImageNette \cite{imagette}. The first row of images shows heatmaps for GradCAM and GradCAM++. The second row of images shows how \sycam  guided by a ground truth metric, captures the ground truth more accurately than the other methods. The last row shows how \sycam guided by the insertion metric generates a heatmap that closely mimics that of the dominant CAM method, ScoreCAM in this case.
    }
    \label{fig:heatmaps}
\end{figure}

In this paper, we address the challenge of incorporating intent by proposing a metric-based approach to generating explanations for CNNs, i.e., where explanations are optimized for a predefined metric. Specifically, we study this problem in the context of explainability methods based on \emph{class activation mappings} (CAM) \cite{CAM}.
CAM methods are one of the most adopted methods for generating saliency maps, i.e.,  heatmaps that highlight the regions of an input most relevant to the CNN's prediction. The definition of relevance differs from one method to another, and thus, each method may result in different heatmaps. \Cref{fig:heatmaps}, shows example heatmaps generated by a set of different CAM methods. In the figure, the heatmaps highlight slightly different image regions across the different techniques. Heatmaps produced by CAM are the result of computing 
a linear combination of the feature maps from the convolutional layers of a CNN. 
CAM methods generate heatmaps by weighting feature maps based on their contribution to the class score. They differ in how they calculate the weights, each offering a unique expression for computing them. Selecting the most appropriate CAM method depends on the specific application and the level of detail in the activation map.

For example, GradCAM \cite{GradCAM} tends to favor larger activation regions in the activation maps, highlighting the most prominent part of an image that influences classification. In an image with multiple swans (see \Cref{fig:heatmaps}, first row), GradCAM primarily highlights the two more visible swans on the right. However, if these swans are removed, the model may still classify the image correctly based on the partial swan. In contrast, GradCAM++ \cite{GradCAM++} incorporates all activation regions, highlighting 
also the partial swans as influential in the CNN’s decision. Depending on the user’s intent, whether they seek to understand the model’s overall behavior or identify the most influential part of a specific image, they may prefer one method over the other.
Choosing the right CAM method is, therefore, often a complex task, traditionally relying on human intuition and empirical experimentation. A non-expert user may not know which CAM expression is most suitable for their application, and a systematic approach to guide the generation of optimal CAM expressions, with respect to certain predefined metrics, is missing.

Our framework, \sycam, presents a metric-based framework for the automatic synthesis of CAM expressions. From a given class of expressions, \sycam can synthesize expressions tailored to specific properties captured via a fitting evaluation metric. 
For instance, metrics capturing the overlap between a heatmap and the ground truth mask \cite{Szczepankiewicz2023} or the pixel intensity of the heatmap in such overlap \cite{schmetric} guide towards the generation of  CAM expressions that generate heatmaps with higher similarity to ground truth. Metrics like the Deletion or Insertion metrics \cite{rise} lead to CAM expressions that generate heatmaps highlighting what the model is truly paying attention to (independent of the correctness of the prediction).
%

Consider the heatmaps shown in the second row in \Cref{fig:heatmaps}, where different saliency methods have been applied to explain why a CNN classified an X-ray image as a COVID case. Here, we used \sycam to synthesize an expression optimized toward the ground truth, i.e., during synthesis, expressions are evaluated based on whether they highlight pixels that also match the ground truth. From the figure, we can see that the heatmap generated according to the \sycam synthesized expression  captured the ground truth more accurately. In contrast, all other methods failed to do so,  producing significantly less accurate representations of the ground truth region.

We applied \sycam using different metrics to study their impact on heatmap generation. In \Cref{fig:heatmaps}, in the last row, we present another example where \sycam was applied using the insertion metric \cite{rise}. The insertion metric evaluates how the highest-scoring pixels in the heatmap influence image classification. In this case, ScoreCAM achieved the highest score compared to GradCAM, GradCAM++, and AblationCAM. \sycam generated an expression that closely resembled ScoreCAM and achieved a very similar score. This demonstrates that when a particular method outperforms others based on a given evaluation metric, \sycam produces expressions that align with the dominant approach. 


Our results are based on an instantiation  of \sycam, adapting techniques from the syntax-guided synthesis literature (SyGuS) \cite{sygus}, which searches a space defined by a grammar of potential CAM expressions. 
This is done via an \emph{oracle-guided inductive synthesis} (OGIS) approach \cite{ogis}. In OGIS, a learner explores the space of possible solutions, guided by oracles that give feedback and evaluate the correctness of solutions generated by the learner. Here, the learner is a synthesis process that searches the space of possible CAM expressions, guided by two oracles: one that can remove equivalent solutions and another that evaluates the candidate solution according to a given evaluation metric. 
We show that, while a monolithic enumerative approach can be used to synthesize CAM expressions, it comes with the limitation of not taking into account any of the image properties, such as the image class. 
To overcome this limitation, we present an adaptation of the synthesis approach leveraging a class-based decomposition of the problem. Specifically, our approach allows for case splits, thereby synthesizing a set of CAM expressions for each class of properties. This is showcased in our experiments, where we use \sycam 
to synthesize CAM expressions for three prominent models: ResNet50, VGG16, and VGG19 \cite{pytorch}, and compare them with those of established methods like GradCAM, GradCAM++, ScoreCAM, and AblationCAM.

%

In summary, \sycam provides a general framework for the systematic generation of CAM methods, with flexibility in two dimensions: first, the user can provide an evaluation metric to suit their use case; and second, the user can provide a syntactic template or grammar that defines the space of possible expressions, giving the potential to provide user intuition to the search algorithm. \sycam will then find an expression 
that is dominant with respect to the given evaluation metric. 
Our primary contributions can be summarized as follows: 
\begin{itemize}
    \item We introduce the problem of synthesizing CAM expressions for CNNs optimized for specific evaluation metrics. 
    \item We present a framework for solving the problem following the OGIS approach, adapting enumerative techniques from SyGuS.
    \item We present a thorough experimental evaluation demonstrating the efficacy of our framework in synthesizing CAM-expressions for the prominent classification models ResNet50, VGG16, and VGG19. 
    \item We evaluate our framework over a COVID-19 benchmark, where the expressions incorporate metrics that favor expert knowledge.
\end{itemize}

\section{Background}

\subsection{Convolutional Neural Networks}
\begin{figure}[t]
    \centering
    \includegraphics[width=.65\linewidth]{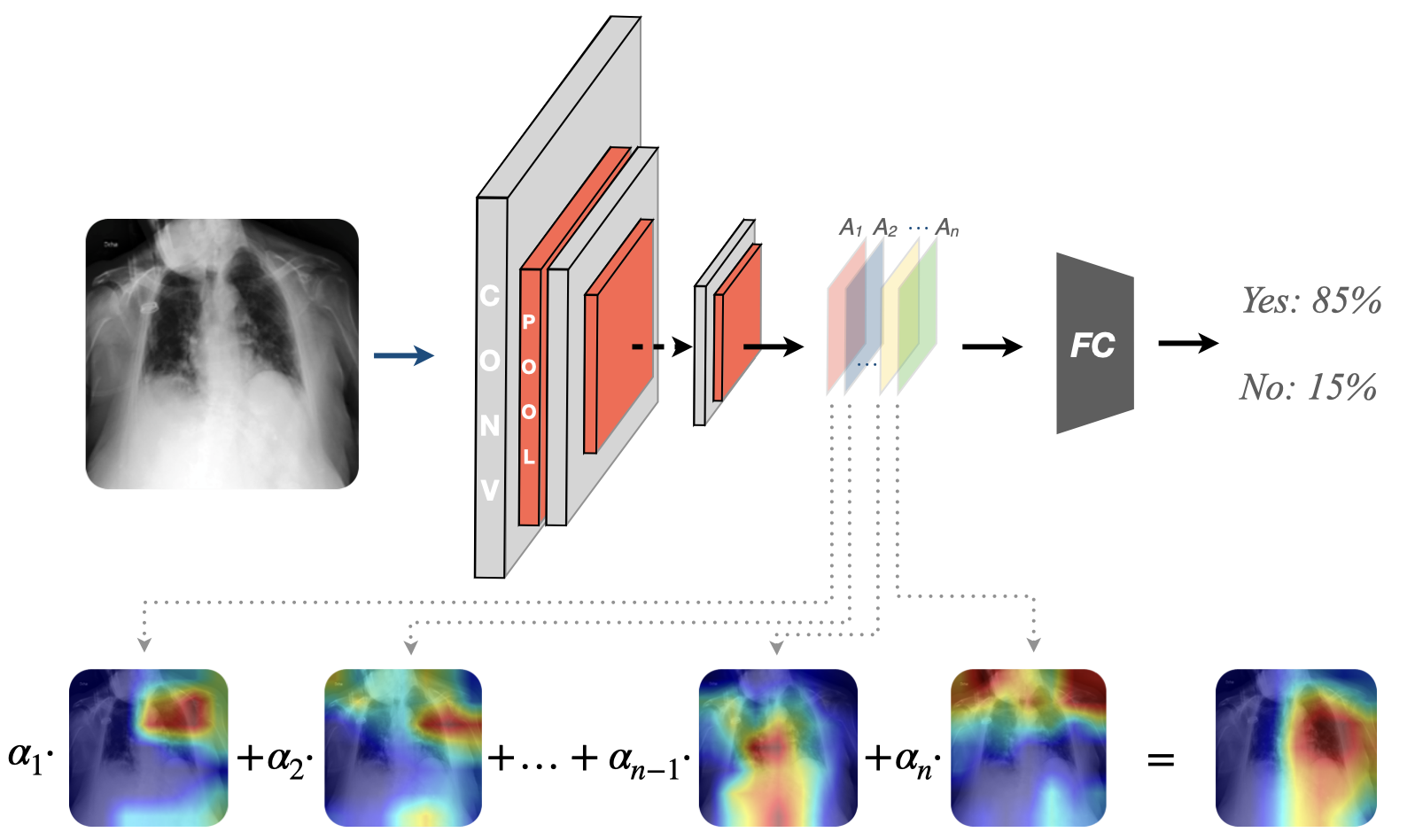}
    \caption{\small Overview of a CNN-based model that classifies X-ray images into COVID-19 positive or negative, and a CAM-based method that explains each classification.}
    \label{fig:convcam}
\end{figure}

Convolutional Neural Networks (CNNs) have proven highly effective in pattern recognition tasks, especially in image processing \cite{6909475,554195,6909619}.
The typical architecture of a CNN is illustrated in \Cref{fig:convcam}. The fundamental building blocks of a CNN are the convolutional layers, which are used for identifying features within an image. These layers work by iteratively applying filters, so-called kernels, to the input image. 
These filters are learned during the training phase. Each layer applies multiple filters to the output of the previous layer, starting with the input image. 
The output produced by applying a filter is known as a feature map. 
As the network deepens, the feature maps define increasingly complex features. While the initial layers may focus on basic elements, such as colors and edges, deeper layers recognize larger patterns. 
In classification tasks, these higher-level features are usually passed to a fully connected layer used for classifying 
the image. To reduce the size of these feature maps while preserving critical information, pooling layers are introduced after each convolutional layer. Common pooling methods include average pooling and max pooling, which reduce a portion of a feature map into its average or maximum value, respectively. 

\subsection{Class Activation Mapping}
 \label{subsection:CAM}

Saliency maps are visual artifacts that highlight the regions of an image most relevant to a model's prediction. Class Activation Mapping (CAM) is a prominent set of methods for generating these saliency maps. The core idea of CAM is to compute a saliency map,  the class activation map, via a weighted linear combination of the feature maps of the last convolutional layer \cite{CAM}. This layer is chosen because it captures high-level features that are most relevant for the classification. The map is always generated for a certain class and points to the parts of an image the network focuses on when predicting that particular class. 

The general concept of CAM is illustrated in \Cref{fig:convcam}.
Formally, for an output class $c$ and a convolutional layer $l$, the saliency map $L^c$ can be defined as: $ L^c = \sum_k \alpha_k^c A_k^l$, 
where $A_k^l$ denotes the $k$-th feature map of the layer $l$,
{i.e., the two-dimensional array resulting from applying the $k$-th convolutional filter of the layer $l$ to its input,}
and the weights $\alpha_k^c$ depend on the CAM method chosen.

Each feature map highlights specific aspects of the image. High values in a feature map point to the presence of a feature; low values hint at its absence. 
Since each feature map contributes differently to the scores of individual classes, the linear combination is weighted to reflect the influence of each feature map in producing the final class activation map. These influences are represented  by the weights $\alpha_1, \dots, \alpha_n$.

Different CAM methods are distinguished by how they calculate the weights. 
In the original CAM, the weights were defined via the learned weights corresponding to different classes. Methods like GradCAM or GradCAM++ \cite{lerma2022gradcam} compute the weights by using the gradients of the score for a specific class with respect to the feature maps of a convolutional layer. Further gradient-based approaches building on the latter two include \textsc{Smooth GradCAM++} \cite{smoothgradpp},
\textsc{Augmented GradCAM} \cite{augmentedGC}, and \textsc{XGradCAM} \cite{XGradCAM}. Gradient-free approaches determine the importance of different regions in the input image for a specific class without relying on gradients. These include perturbation-based approaches like \textsc{ScoreCAM} \cite{scoreCAM} and \textsc{AblationCAM} \cite{ablationcam}, attention-based methods like Attention-Guided {CAM} \cite{attentioncam}, and methods like \textsc{EigenCAM} \cite{eigenCAM}, that apply principal component analysis to create the class activation map.  For a comprehensive survey, we refer the reader to \cite{survey}.
So far, no systematic approach has been used to determine the optimal weight expression for a given task. Our work addresses this challenge by introducing an automated approach capable of synthesizing both gradient-based and gradient-free expressions if the grammar allows it.

\subsection{Heatmaps Evaluation Metrics} \label{subsection:evaluationmetric}
Several metrics have been introduced in the literature to evaluate saliency methods \cite{XAItaxonomy}. Some of these methods are perturbation-based, i.e., they evaluate the effect of masking regions highlighted by CAM methods on the model's performance. Examples of such metrics include Average Drop \% \cite{GradCAM++}, AOPC \cite{AOPC}, ROAD \cite{road}, IROF \cite{IROF}, and the Deletion and Insertion \cite{rise} metrics.
Another category of metrics is ground-truth-based. These metrics measure the distance of the explanation to the ground-truth explanation data. Some examples of these metrics are the $m_{GT}$ metric defined in \cite{Szczepankiewicz2023}, the Segmentation Content Heatmap (SCH) metric \cite{schmetric}, CEM \cite{cem}, and CLEVR-AI \cite{clevr}. 
\sycam is agnostic to the chosen metric, and CAM-weight expressions can be synthesized by \sycam with respect to any of the aforementioned metrics. Each evaluation metric focuses on different properties of the heatmaps, e.g., similarity to ground truth, robustness, and more. The synthesized expressions are, in consequence, based on well-founded criteria, guided by quantitative evaluation metrics. The resulting expressions generate optimized saliency maps according to these metrics, thus allowing us to eliminate any human biases and to select a CAM method more suitable for a given context. An extensive study on the correctness of XAI techniques in generating the true explanations, the so-called fidelity, can be found in \cite{fidelity}. 

In the following, we provide a detailed introduction to key evaluation metrics that we also use later in our experiments.

\paragraph{Average Drop \%:}
The intuition behind this metric is one that checks whether removing parts not highlighted by the heatmap of an image reduces the classification confidence of the model. Removing parts of an image that are not relevant shouldn't heavily impact the confidence drop. 
To check if the most relevant parts of an image $i$ are preserved, the product of the heatmap and the image is computed. Then, the resulting image $h$ is classified to measure the confidence drop. A low confidence drop implies that the heatmap contains the most relevant features of the image, so the lower the metric value, the better.

Given a dataset $\mathcal{I}$, the Average Drop \% is expressed as $\sum_{i \in \mathcal{I}} \frac{\max(0, y_i^c - h_i^c)}{y^c_i} \cdot 100$, where $y_i^c$ is the classification score for image $i \in \mathcal{I}$ and class $c$, and $h_i^c$ is the classification score for the product of image $i$ and the generated heatmap for such image, and class $c$. 

\paragraph{Deletion and Insertion metrics:} 
The Deletion metric measures the drop in the classification score when the most relevant pixels of the image are \emph{gradually} removed, while the Insertion metric measures the rise in the classification score when they are iteratively added to a blank image. In this paper, we use modified versions of the Deletion and the Insertion metrics.
Given a model $M$, an image $x$, a limit on the number of perturbations $P$, and a saliency map $\mathcal{H}$ over $x$, the process recursively modifies the image according to the following formula:
\begin{align*}
         & x^{(0)} = x \quad \quad \quad 
         \forall \ 1 \leq j \leq P \ : \  x^{(j)} = g(x^{(j-1)},\overline{x}, r_j)
\end{align*}
For the Deletion metric, the function $g$ gradually replaces the most relevant pixels $r_j$, ordered by the saliency map $L^c_M[e]$, and their neighborhoods with the corresponding parts of a highly blurred version of $x$, denoted by $\overline{x}$. The neighborhood of a pixel is given by the feature maps used to compute the saliency map. If the feature maps $A_k$ have size $w \times h$, we split the image into a grid of $w \times h$ neighborhoods. For the Insertion metric, the process is analogous, but the most important pixels are added to the highly blurred version of the image instead. The metrics are then defined by: 
\[ \mu_\textit{deletion} = \frac{1}{|I|} \sum\limits_{x \in I} \frac{1}{P+1} \Big( \sum\limits_{j=0}^P M(x^{(0)})^{c_0} - M(x^{(j)})^{c_0} \Big)  \]
\[ \mu_\textit{insertion} = \frac{1}{|I|} \sum\limits_{x \in I} \frac{1}{P+1} \Big( \sum\limits_{j=0}^P M(x^{(j)})^{c_0} - M(x^{(0)})^{c_0} \Big)  \]
where $c_0$ is the output class of $M(x^{(0)})$, $M(x)^{c_0}$ is the output score for class $c_0$, and $|I|$ is the number of images in the set $I$.
Unlike for other similar metrics \cite{artifacts}, no artifacts are added through this procedure, and $\overline{x}$ is generated using deterministic filters, so the influence of random perturbations over the metric is also avoided. The purpose of using these versions instead of the originals \cite{rise} is to avoid adding artifacts, to make sure that the computed scores are relative to the initial score of the model $M$ over the base image, and to make them work so that the higher the values, the better. 

Intuitively, higher values of the metrics imply higher variations in the classification score during the perturbation process, i.e., higher relevance of the pixels highlighted by the saliency map. 
We may get some insights about the datasets employed using these metrics. For example, when the confidence drop for the Deletion metric is much lower for a subset than for the others, it means that the model is still confident in its classification even after masking portions of the objects. This may imply that the objects occupy most of the image or that the model identifies the objects through the background. In such cases, it is important to improve the dataset.

\paragraph{Ground truth similarity metrics:} Some evaluation metrics are focused on measuring the similarity of the heatmap to a ground-truth mask. 

Given a heatmap $H$ and a ground-truth mask $H_{GT}$ with $p$ pixels, the $m_{GT}$ metric takes the $p$ most relevant pixels of $H$ and counts how many of those pixels are part of the ground-truth mask. This is, if $n$ of the $p$ most relevant pixels of $H$ are part of the mask $H_{GT}$, then $m_{GT}(H, H_{GT}) = n/p$. While the $m_{GT}$ metric measures the similarity between a heatmap and the ground truth, it doesn't provide information about the intensity of the heatmap pixels. The SCH metric, on the other hand, solves this problem. The SCH metric is given by:
\begin{align*}
    \mathit{SCH}(H,H_{GT}) = \frac{\sum\limits_{i,j} H_{i,j} \cdot M_{i,j}}{\sum\limits_{i,j} H_{i,j}}
\end{align*}
Intuitively, the \textit{SCH} metric measures how concentrated the heatmap is in the ground-truth mask. The more concentrated the pixels are and the higher their relevance, the higher the metric will be, so the higher, the better.

Notice that a strong assumption of ground-truth similarity metrics is that the models are making classifications based on the ground-truth part of the image. If the models are not doing so, and are making decisions based on other information found in the input data, these metrics would return low values independently of how well the heatmaps are explaining what the models are paying attention to. 


%

%
\subsection{Syntax-Guided Synthesis}
\label{subsection:progsynt}
Syntax-Guided Synthesis (SyGuS)~\cite{sygus} is the problem of generating a function that both satisfies a semantic specification and is contained within a language described in a context-free grammar (CFG). 
\begin{definition}[Context-Free Grammar]
A context-free grammar $G$ is defined as 
 as a set of terminal symbols $T$ and a set of nonterminal symbols $NT$, a start symbol $S$, and a set of production rules $R \subseteq NT \times (T \cup NT)^*$ which describe the way programs may be constructed iteratively from the grammar. 
 \end{definition}
 \begin{definition}[SyGuS problems]
Formally, a SyGuS problem is a 
4-tuple $\langle T,G,\phi,F\rangle$ such that $T$
is a first-order theory, $G$ is a context-free grammar, $\phi$ is a first-order
formula, and $F$ is a function symbol that may occur in $\phi$.
A solution to a SyGuS problem
$\langle T,G,\phi,F\rangle$ is a function $f$ such that
$T \models \phi[F \mapsto f]$ and $f \in \mathcal{L}(G)$, where $\phi[F \mapsto f]$ denotes replacing all occurrences of $F$ in $\phi$ with $f$. 
\end{definition}

Later in the paper, we study a  SyGuS instantiation of CAM expression synthesis, where the specification is that the resulting CAM method must perform better than a given threshold function. This is described in detail in Section~\ref{sec:implementation}.
The most common approach for solving SyGuS problems is Oracle Guided Inductive Synthesis (OGIS)~\cite{ogis}; a family of algorithms that alternate between a learner, which attempts to learn a solution to the synthesis problem, and an oracle, which guides the learner via means of queries and responses, the simplest of which is a correctness query (the learner asks ``is this candidate program correct?'', and the oracle replies with ``yes'' or ``no''). There is a broad variety of learners in the literature, but the most common are enumerative techniques. 
We take inspiration from some of the most common enumerative learners~\cite{bottomupsearch} when implementing our approach. 

\section{CAM Expression Synthesis}
\label{section:problem}

\subsection{Problem Statement} \label{section:problemstatement}
Let $t[?]$ define an expression with holes, known as a \emph{template}, where holes in the expression are marked by the symbol $?$. 
For an expression $e$, the expression $t[e]$ results from replacing every appearance of the symbol $?$ in $t$ with $e$. 
Following the definition of CAM as given in  \Cref{subsection:CAM}, we define a CAM template as $L^c[?] = \sum_{k} ?\cdot A_k^l$. The $?$ is a placeholder for an expression that defines how the weights $\alpha^c_k$ are computed for a class $c$. We refer to such expressions as CAM-weight expressions. We refer to the set of all instantiations of $L^c[?]$ by $\cal F_{\textit{CAM}}$.

\begin{problembox}[CAM expression synthesis]
Let $\cal M= (\mathcal{I} \rightarrow \mathbb{R}^{|C|})$ be a set of CNN-based classifiers defined over a space of images $\cal I$ and a set of classes $\cal C$.
Given $M\in \cal M$,  
a set of images $I\subseteq \cal I$, 
a threshold function $\lambda\colon \cal M\times \cal I \rightarrow \mathbb R$, 
a set of CAM-weight expressions $\cal E$, 
and an evaluation function $\mu\colon \cal F_{\text{CAM}} \times \cal M \times \cal I \rightarrow \mathbb R$, synthesize an expression $e\in \mathcal{E}$ s.t. $\mu(L^c[e], M,I)> \lambda(M,I)$.
\end{problembox}

In our problem statement, the role of the threshold function $\lambda$ is to set a lower bound on the quality of synthesized expressions with respect to the evaluation function $\mu$. As we will see in our experiments, the threshold function can be given as a fixed number or as a function of any other CAM function. The evaluation function $\mu$ defines a metric for evaluating CAM functions and can be realized by implementing known CAM evaluation metrics from the literature. 

\subsection{Oracle-Guided Synthesis of CAM-weight expressions}
\label{section:framework}

Now that we have a formally defined problem statement, we can discuss the \sycam framework, an oracle-guided synthesis approach for the synthesis of CAM-weight expressions. Let us initially assume we have access to two oracles:
\begin{itemize}
\item an equivalence oracle: this oracle receives as input a set of expressions $\epsilon$, which may contain semantically equivalent expressions, and returns a set of expressions $\epsilon' \subseteq \epsilon$ such that no two expressions in $\epsilon'$ are semantically equivalent; and
\item a correctness oracle: this oracle receives a single CAM-weight expression as input, and returns a boolean which is true if the CAM-weight expression results in a saliency map that scores above a pre-defined threshold on a pre-defined metric, for a given set of images and classification model. 
\end{itemize}

We will define the \sycam framework assuming access to these oracles, but we should note that these oracles are performing tasks that are, in general, undecidable (equivalence checking) or at least computationally expensive (evaluating a given CAM weight expression across a large set of images). We will address the practical implementation of each of these oracles in Section~\ref{sec:implementation}. 

The general workflow of \sycam is depicted below. \sycam is composed of two main procedures, the synthesis phase (which is guided by the equivalence oracle) and the evaluation phase (the correctness oracle). Candidate expressions produced by the synthesis procedure 
are generated from a space of expressions defined by an input grammar $G$, as described in \Cref{sec:implementation}.
The expressions are then forwarded one by one to the evaluation process. An expression is evaluated using a correctness oracle, defined in terms of the given evaluation metric $\mu$ and a threshold function $r$, over a set of images $I$. If a candidate passes the evaluation process, it is returned as a solution to the overall synthesis process. If the evaluation fails, another expression is selected out of the current list of candidate expressions, and the same evaluation process is repeated for the new expression. If all candidate expressions have been evaluated with no success, the synthesizer is triggered again to generate a new set of candidates. 

\begin{center}
    \scalebox{0.8}{\def\checkmark{\tikz\fill[scale=0.1](0,.35) -- (.25,0) -- (1,.7) -- (.25,.15) -- cycle;} 
\begin{tikzpicture}[>=latex,x=3cm,y=2cm]
\node[rectangle,draw,minimum height=1.2cm,minimum width=2.5cm] at (1,1) (s) {\textsc{synthesize}};
\node[rectangle,draw,minimum height=1.2cm,minimum width=2.5cm,align=center] at (1,0) (e) {\textsc{Equivalence}\\ \textsc{Oracle}};
\node[rectangle,draw,minimum height=1.2cm,minimum width=2.5cm,align=center] at (2.2,1) (v) {\textsc{Correctness}\\ \textsc{Oracle}};

\node[anchor=west] at (2.9,1) (sol) {solution};
\node[anchor=west] at (0.1,1) (grammar) {$G$};

\path[->] ($(s.east)+(0,+0.1)$) edge[above] node {$\epsilon$} ($(v.west)+(0,0.1)$);
\path[->] ($(v.west)+(0,-0.1)$) edge[below] node {$\times$} ($(s.east)-(0,0.1)$);
\path[->] (v.east) edge[above] node {\checkmark} (sol);
\path[->] ($(s.south)+(-0.1,0)$) edge[left] node {$\epsilon$} ($(e.north)+(-0.1,0)$);
\path[->] ($(e.north)+(0.1,0)$) edge[right] node {$\epsilon'$} ($(s.south)+(0.1,0)$);
\path[->] ($(grammar.east)+(0,0)$) edge[above] node {} ($(s.west)+(0,0)$);

\end{tikzpicture}
}
\end{center}

\paragraph{Synthesis phase:} The task of the synthesis phase is to enumerate expressions using the feedback given by the oracles. 
The algorithm we use is based on a classical program synthesis technique: the bottom-up search algorithm \cite{bottomupsearch}. For the algorithm, we require an input grammar $G$, which defines an initial set of expressions (\textit{terminals}) and production rules $R$ that allow us to combine expressions to synthesize new ones (\textit{synthesize} block). The synthesis phase calls the equivalence oracle, \textsc{elimEquiv}, in order to reduce the exponential growth in expressions.
This bottom-up search algorithm is shown in Algorithm~\ref{alg:bottomupsearch}.

\begin{algorithm}[t]
\begin{multicols}{2}
   \begin{algorithmic}
    \Function{Search}{$G$, $\lambda$, ${M}$, $I$, $\mu$}
        \State $\mathit{exprs}$ $\gets \emptyset$
        \State $\mathit{exprs'}$ $\gets$ $G.\mathit{Term}$ 
        \While{True}
        \State{$solution \gets \Call{Eval}{\mathit{exprs'}, \lambda, {M}, {I}, \mu}$}
        \If{$solution \neq \text{nil}$}
        \State \textbf{return} $solution$
        \EndIf
        \State $\mathit{exprs} \gets \mathit{exprs} \cup \mathit{exprs'}$
        \State $\mathit{exprs'} \gets \Call{Expand}{\mathit{expr},G}$
        \State $\mathit{exprs'} \gets \Call{elimEquiv}{\mathit{exprs}, \mathit{exprs'},{I}}$
        \EndWhile
    \EndFunction 
   
    
    \Function{Expand}{$\mathit{E,G}$}
    \State $E' \gets \emptyset$
        \For{$r \in G.R$}
        \If {$r$ is $1$-ary}
        \For{$e_1 \in E$}
        \State $E' \gets E' \cup r\{e_1\}$
        \EndFor
        \ElsIf{$r$ is $2$-ary}
        \For{$e_1, e_2 \in \Call{Prod}{E,E}$}
        \State $E' \gets E' \cup r\{e_1, e_2\}$
        \EndFor
        \EndIf
        \EndFor
        \State \textbf{return} $E'$
    \EndFunction
    \end{algorithmic}
\end{multicols}
    \caption{Bottom-up Search}\label{alg:bottomupsearch}

\end{algorithm}

Initially, we populate \textit{exprs} with all expressions for \textit{Term}. At each iteration of the Algorithm \ref{alg:bottomupsearch}, the search process deploys the \textsc{Expand} function to iterate through the production rules of the grammar, and generates all possible new expressions that use the elements in \textit{exprs} to replace the non-terminals in each production rule and adds them to \textit{exprs}. We use $r\{e_1, \ldots, e_n\}$ to indicate the result of taking the rule $r \in R$ and replacing the first nonterminal symbol occurring in $r$ with $e_1$, the second with $e_2$, and so on. 
\paragraph{Oracles:} The equivalence oracle \textsc{ElimEquiv} then reduces the set of expressions, by removing all semantically equivalent expressions. At each iteration, the correctness oracle \textsc{Eval} checks the current list of programs to see if it contains a program that it deems correct (i.e., a program that performs above a given threshold on the evaluation metric). If so, the program is returned as a solution.

\section{An Instantiation of SyCAM}
\label{sec:implementation}

Our approach is customizable to any grammar. In this section, we give an instantiation of our framework for a grammar defining gradient-based expressions, and expressions based on ScoreCAM and AblationCAM.  We use a grammar with non-terminals $NT = \{Expr, Term, Grads\}$, and starting category $S = \{Expr\}$.
The set of production rules $R$ is defined as follows:
{\small
\begin{align*} 
\it {Expr} := \ & \it{Term} \ |  \ 
{Expr} + {Expr} \ | \
2 \cdot {Expr} + {Expr} \ | \  \it {Expr} \cdot {Expr} \ | \ \mathit{ReLU}(Expr) \\
\mathit {Term} := \ & \mathit{Grads} \ | \ \mathit{top}_5(\mathit{Grads}) \ | \ \mathit{top}_{10}(\mathit{Grads}) \ | \ \mathit{top}_{20}(\mathit{Grads}) \ | \ \mathit{top}_{50}(\mathit{Grads}) \\
& \mathit \ | \  \mathit{CICScores} \ | \  \mathit{AblScores}  \\
\mathit {Grads} := \ & \mathit GP(\frac{\partial Y^c}{A_k}) \text{ for any } c \in C \text{ and } k \in K.
\end{align*}
}
where $\mathit{ReLU}$ is the element-wise function $\max\{\cdot, 0\}$. 
Inspired by algorithms like GradCAM, we use as terminals the \textit{gradients}, denoted by \textit{Grads}, of the score for the predicted class $c$ with $Y^c = \mathcal{M}(x)$ with respect to the feature map activations $A_k$ of the last convolutional layer, i.e. $GP(\frac{\partial Y^c}{A_k})$ where $GP$ denotes the global average pooling operation, for each feature map $k$ in the set of feature maps $K$ and for each class $c$ in the set of classes $C$. We also include as terminals the functions $\mathit{top}_n(\mathit{Grads})$ that nullify all but the highest $n$ elements of \textit{Grads} and are denoted by $\mathit{top}_n$. 

Inspired by ScoreCAM, we include as terminals ($\mathit{CICScores}$) the \textit{channel-wise Increase of Confidence (CIC)} \cite{scoreCAM}. For each feature map $k$, each image $x$, a baseline input image $x_b$, and a model $\mathcal{M}$, the CIC is defined as $CIC(A^k) = \mathcal{M}(x \circ H^k) - \mathcal{M}(x_b)$ where $H^k = s(Up(A^k))$, $Up$ denotes the upsample operation that upsamples $A^k$ into the input image size, and $s(\cdot)$ is a normalization function to the range $[0,1]$. We assume that $\exists x_b : \mathcal{M}(x_b) = 0$, and use $C(A^k) = \mathcal{M}(x \circ H^k)$.

$\mathit{AblScores}$ represent the weights used in the definition of AblationCAM \cite{ablationcam}. For each feature map $k$ of a model $\mathcal{M}$, and each image $x$, these weights $w$ are defined by $w_k^c = \frac{y^c-y^c_k}{y^c}$, where $y^c = \mathcal{M}(x)$ and $y^c_k$ is the result of setting all the activation cell values of $A_k$ to zero and classifying again the $x$. 

Notice that the terminals of the grammar are vectors with number of elements equal to feature maps in the last convolutional layer of $\mathcal{M}$, and every expression generated by this grammar produces vectors with the same size as the terminals. 

\subsection{Equivalence Oracle} \label{sec:equivalence}
In general, determining the equivalence between two expressions is undecidable. We thus use an approximation, referred to as observational equivalence. 
\begin{definition}[Observational Equivalence]
Formally, two expression $e_1$ and $e_2$ are observational equivalent on a finite set of images $I$, according to an evaluation metric $\mu$ and a model $M$ iff $\mu(L^c[e_1], M,I) = \mu(L^c[e_2], M,I)$.
\end{definition}

If two expressions are observationally equivalent, we can remove one of these expressions from the pool of expressions used to build subsequent programs \emph{provided} the semantics of the program fragment do not depend on context. 
That is, given two expressions $e_1$ and $e_2$, and a set of images $I$, if $e_1$ and $e_2$ give the same result on the set of images, then so will $C[e_1]$ and $C[e_2]$ where $C[e_1]$ is a program that uses $e_1$ as a subexpression, and $C[e_2]$ is the same program but with $e_2$ in place of $e_1$.

CAM-weight expressions are arithmetic expressions so this property is true, \emph{for the set of images $I$}. However, since evaluating the expression on all the images in our dataset is time-consuming and impractical, our observational equivalence oracle uses a smaller subset of the full dataset. Thus, if this set is not representative of the full dataset, the observational equivalence oracle may remove expressions that we subsequently may need. As a consequence of this approximation of equivalence, SyCAM may fail to find some possible solutions, but this trade-off is worth it to prune the exponential growth of the search space.

\subsection{Correctness Oracle} \label{section:correctness}

Recall that we wish to synthesize an expression $e$ such that the following is true;
$\mu(L^c[e], M,I)> \lambda(M,I)$, for a given $I$ and $M$. 
Unlike many domains where OGIS is used, it is not possible for us to reason about this expression symbolically using techniques like SMT solvers, since this would require us to reason about the weights of the model and the pixels of each image. We must therefore use testing, executing each generated CAM method on the images in the set $I$. Expressions are then checked one by one for their correctness. If an expression passes the correctness test, the overall synthesis process terminates, returning this expression as a solution. As we will see in the experiments, this process is modified to run for a fixed amount of time and then return the best expression synthesized instead.

\subsection{Class-based decomposition}
\label{section:divide}

A limitation of the algorithm presented in the previous section is that the grammar we use does not contain any expressions that can perform case splits, e.g., an ``if'' expression, or any logical expressions that can define when to apply a specific CAM method to a particular image. This means that the function we synthesize is applied uniformly to all images, regardless of the properties of that image. In this section, our goal is to extend this grammar to permit case splits. One obvious way of doing this would be to introduce an ``if-then-else'' statement into the grammar $G$, as well as expressions for identifying features or characteristics of different images. This, however, results in a significantly expanded search space, and an intractable synthesis problem. 
In this section, thus, we break down the synthesis method into two parts: a \textit{classification} model $\mathcal{M}$, and a set of CAM-weight expressions, synthesized using the enumeration approach described previously, that should be applied to each class. Thus, for a given set of classes, the end CAM-weight expression will be an expression in the grammar given by $G$ extended with the following production rule:
$$
\mathit{Expr} := (Y^{c_i} = max(Y^{c_1}, \ldots Y^{c_n})) \,? \,\mathit{Expr} \, : \, \mathit{Expr}
$$
where $Y^{c_i}$ is the confidence score for class $c_i$, and $\{c_1, \ldots c_n\}$ is the set of classes generated by the model $\mathcal{M}$.
%
%
It would be possible to use any classifier in this step, but we take advantage of having a classifier that can choose the expression for each image: the model $\mathcal{M}$. 

One advantage of this approach is that it prevents the rejection of good expressions that do not perform well only for small subsets of images. 
%
However, an important disadvantage is that the algorithm has to be executed once for each output class. If the number of output classes of a model is in the order of thousands, this method requires considerable computing power.

\section{Experiments}\label{section:experiments}
\label{sec:experiments}





In this section, we present three sets of experiments that show the efficacy of \sycam: 
\begin{itemize}
    \item \textbf{E1 - Sanity check:}  We show the ability of \sycam to synthesize known dominant CAM expressions for a given metric, or better ones if the grammar allows for more expressive expressions. We apply \sycam to one Pytorch model, VGG16 trained over the PASCAL VOC 2007 dataset \cite{pascal-voc-2007}.
    \item \textbf{E2 - Enumeration vs class-based decomposition:}  We compare \sycam to the classical approaches of GradCAM, GradCAM++, ScoreCAM, and AblationCAM. Here, we particularly compare the enumerative and class-based decomposition approaches and show how the latter improves over the former. \sycam is applied to three PyTorch models: ResNet50, VGG16, and VGG19, trained over the Imagenette dataset \cite{imagette}.
    \item \textbf{E3 - Incorporating ground truth:} This experiment is an application of \sycam to show that, in contrast to standard methods, \sycam allows us to incorporate expert knowledge into the generation of saliency maps, resulting in better saliency maps. \sycam is applied to a ResNet50 model trained over COVID-19  X-ray images from the COVID-QU-Ex dataset ~\cite{covid_dataset}. 
\end{itemize}  


  All experiments were run using an NVIDIA T4 with 16GB RAM. The code used can be found in \url{https://github.com/starlab-systems/SyCAM}. The computations were enabled by resources provided by the National Academic Infrastructure for Supercomputing in Sweden (NAISS), partially funded by the Swedish Research Council through grant agreement no. 2022-06725. 




\subsection{Experimental setup} \label{section:experimentalsetup}


As established in \Cref{section:problemstatement}, the goal is to find an expression  $e$ such that $\mu(L^c[e], M,I)> \lambda(M,I)$, for a given image set $I$ and a model $M$ by following  \Cref{alg:bottomupsearch}. We adapt \Cref{alg:bottomupsearch}, to one that instead of stopping the computation the moment we find an expression $e$ that beats a threshold $\lambda$, to one that continues the search, always saving the so-far best expression found, and taking the value $\mu(L^c[e], M,I)$ for the best expression as the new threshold $\lambda$. We start with $\lambda=0$. We let the experiments run for a fixed amount of time and return the best expression synthesized during that time.

\paragraph{Grammar:} The grammar used for the experiments is defined in \Cref{sec:implementation}. Computing the weights generated by terminals $\mathit{CICScores}$ and $\mathit{AblScores}$ is computationally expensive. In the case of ResNet50, VGG16, and VGG19, they are required to compute 512 classifications per image. To reduce the computation overhead, for each dataset and evaluation metric, we precompute the weights produced by these terminals 
beforehand for each image in the dataset. 

\paragraph{Synthesis phase:} The set of expressions generated grows exponentially. Because we have 7 terminals in the grammar, and we are enumerating solutions that can be produced with the rules defined, more than $1000$ expressions were generated in only three applications of the \textsc{Expand} function (see \Cref{alg:bottomupsearch}) even after discarding equivalent expressions using observational equivalence (\Cref{sec:equivalence}). By the fourth application of the \textsc{Expand} function, we would generate more than $10^6$ expressions. To prevent spending too much time just discarding equivalent expressions without evaluating them, instead of generating a whole new set of expressions with \textsc{Expand} before evaluating, we generate and yield expressions one by one. Each expression is evaluated just after being generated by the equivalence oracle and, if not discarded, by the correctness oracle.

\paragraph{Equivalence oracle:} The function \textsc{ElimEquiv} of \Cref{alg:bottomupsearch} employs a subset with 10 images, one of each class of the dataset, to ensure this subset is as representative as possible. We also tested subsets of size 20 and 30 of the Imagenette dataset, and we confirmed that the number of expressions discarded decreases (by 1.5\% and 7.2\%, respectively) with the number of images in the subset. However, the expressions synthesized were the same, and there is a trade-off with the computation time dedicated to generating expressions and discarding those equivalents, so we maintained the original size of 10 images.

\noindent \textit{Correctness oracle:} It would be time-consuming to test every generated CAM method on every image in the dataset $I$. For example, it takes around 18 minutes to compute the Deletion metric for ResNet50 and a single expression over a dataset of 4000 images. To overcome this computation overhead, we implement the correctness oracle as follows, aiming to discard expressions as early as possible.

The evaluation is done by applying an evaluation procedure on the candidate expression defined in terms of an evaluation metric $\mu$ and over increasingly large sets of images $I_1 \subset I_2 \subset \ldots \subset I$. If a candidate expression $e$ is evaluated over the set $I_i$ and results in a score larger than that computed by a threshold function $\lambda$ over at least half of the images in $I_i$, and the average score is better, i.e. $\mu(L^c[e], M, I_i) > \lambda(M, I_i)$, it is then evaluated over the next largest set of images, $I_{i+1}$. Otherwise, the candidate is discarded. If a candidate expression $e$ is evaluated on the set $I$ and achieves a score higher than that computed by the threshold function over at least half of the images, and the average score is higher, then $e$ is the best expression found, and $\lambda$ is updated with the new threshold defined by $e$. 
With this multi-layered approach, we can quickly eliminate programs that already fail evaluation on smaller sets of data, thus speeding up the search for solutions. It is possible that a candidate may perform poorly on a subset $I_i$ and then perform better than the threshold function on a subset $I_{i+1}$, and thus be discarded early. This risk is small since our subsets are relatively large and are uniformly sampled from the full set of images, so likely to be representative. 

The oracle checks if the candidate expression $e$ is better than the best solution found for at least half of the images because the goal is to find an expression that works well for as many images as possible, i.e., that generalizes well, and at the same time has a higher score. 

Lastly, we set a timeout of 6 hours for experiment E1 and 24 hours for E2 and E3, i.e., the returned expression is the best expression found within these time bounds. We note that within this timeout, we were able to reach the fourth iteration of the algorithm, but despite the optimizations, we are not able to cover all the $10^6$ expressions of that iteration. This results in a limitation on the size of the expressions that can be synthesized. However, we show later that \sycam manages to synthesize better expressions than widely known CAM methods. Furthermore, once the \sycam expression is synthesized, the computational cost of generating saliency maps is similar to that of other CAM expressions, so it is reasonable to spend time on generating the best possible expression. The cost depends on the terminals used. For example, the cost of generating saliency maps for an expression that computes the $\mathit{CICScores}$ as part of the expression is similar to the cost of ScoreCAM.

\subsection{Experiments E1: Sanity Check}
\label{section:sanitycheck}
\textit{"Show the ability of \sycam to synthesize known CAM expressions if they are dominant for a given metric, or even better ones if the grammar allows for more expressive expressions"}

In this experiment, we applied \sycam to a VGG16 model trained over the training subset of the PASCAL VOC 2007 dataset, and using the Average Drop \% metric: one of the settings described in the original GradCAM++ paper \cite{GradCAM++}. The dataset $I$ contains 2510 images distributed in 20 classes. For the correctness oracle, we use three subsets of images, $I \supset I_2 \supset I_1$. The set ${I}_2$ is a subset of 1000 randomly chosen images, and ${I}_1$ contains 100 randomly chosen images. 

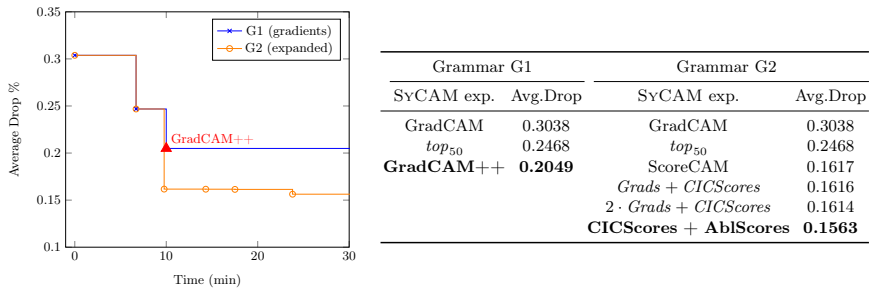
\begin{figure*}[t]
    \begin{minipage}{0.4\textwidth}
    \scalebox{0.55}{
    \begin{tikzpicture}
    \begin{axis}[
            xlabel={Time (min)},
            ylabel={Average Drop \% },
            xmin=-1, xmax=30,
            ymin=0.1, ymax=0.35,
            xtick={0,10,20,30},
            ytick={0,0.05,0.1,0.15,0.2,0.25,0.3,0.35},
            legend pos=north east,
            ymajorgrids=false,
        ]
        \addlegendimage{color=blue, mark=x};
        \addlegendimage{color=orange, mark=o};
        \addplot[color=blue, mark=None] coordinates {(0,0.3038)(6.7,0.3038)(6.7,0.2468)(10,0.2468)(10,0.2049)(2000,0.2049)};
        \addplot[color=blue, mark=x, only marks] coordinates {(0,0.3038)(6.7,0.2468)(10,0.2049)};
        \addplot[color=orange, mark=None] coordinates {(0,0.3038)(6.7,0.3038)(6.7,0.2468)(9.77,0.2468)(9.77,0.1617)(14.32,0.1617)(14.32,0.1616)(17.51,0.1616)(17.51,0.1614)(23.81,0.1614)(23.81,0.1563)(2000,0.1563)};
        \addplot[color=orange, mark=o, only marks] coordinates {(0,0.3038)(6.7,0.2468)(9.77,0.1617)(14.32,0.1616)(17.51,0.1614)(23.81,0.1563)}; 
        \addplot [color=red, mark = triangle*, mark size=4pt] coordinates {(10,0.2049)};
        \addplot [color=red, mark = none, nodes near coords=GradCAM++,every node near coord/.style={anchor=180}] coordinates {(10,0.215)};
        \legend{G1 (gradients),G2 (expanded),}
    \end{axis}

\end{tikzpicture} 
}
\end{minipage} 
\begin{minipage}{0.45\textwidth}
\scalebox{0.7}{
\begin{tabular}[t]{c c c c}
        \toprule
        \multicolumn{2}{c}{Grammar G1} & \multicolumn{2}{c}{Grammar G2} \\
        \cmidrule(l){1-2} \cmidrule(l){3-4}
        \sycam exp. & Avg.Drop & \sycam exp. & Avg.Drop  \\
        \midrule
        GradCAM & 0.3038 & GradCAM & 0.3038  \\
        $\mathit{top}_{50}$ & 0.2468 & $\mathit{top}_{50}$ & 0.2468 \\
        \textbf{GradCAM++} & \textbf{0.2049} & ScoreCAM & 0.1617 \\
        & & $\mathit{Grads} + \mathit{CICScores}$ & 0.1616 \\
        & & $2 \cdot \mathit{Grads} + \mathit{CICScores}$ & 0.1614 \\
        & & \textbf{CICScores} \textbf{+} \textbf{AblScores} & \textbf{0.1563 }\\
        \bottomrule
\end{tabular}}
\end{minipage} 
\caption{\sycam application to a VGG16 model trained over the PASCAL VOC 2007 dataset and the Average Drop \% metric (lower is better). If only gradients-related terminals are included in the grammar, \sycam synthesizes GradCAM++. Better CAM expressions are synthesized for an expanded grammar.}
\label{fig:sanitycheck}
\end{figure*}

We applied \sycam using two grammars for a fixed time of 6 hours each:
\begin{itemize}
    \item G1: A grammar defined by:
    {\small
    \begin{align*} 
    \it {Expr} := \ & \it{Term} \ |  \ 
    {Expr} + {Expr} \ | \
    2 \cdot {Expr} + {Expr} \ | \  \it {Expr} \cdot {Expr} \ | \ {ReLU}(Expr) \\
    \it {Term} := \ & \mathit{Grads} \ | \ \mathit{top}_5(\mathit{Grads}) \ | \ \mathit{top}_{10}(\mathit{Grads}) \ | \ \mathit{top}_{20}(\mathit{Grads}) \ | \ \mathit{top}_{50}(\mathit{Grads})  \\
    \mathit{Grads} := \ & \it GP(\frac{\partial Y^c}{A_k}) \text{ for any } c \in C \text{ and } k \in K.
    \end{align*}
    }
    \item G2: The grammar G1 adding the terminals $\mathit{CICScores}$ and $\mathit{AblScores}$. This is, the grammar described in \Cref{sec:implementation}.
\end{itemize} 
Using grammar G1, \sycam can only synthesize expressions that employ gradients, so \sycam can't synthesize ScoreCAM, AblationCAM, or other similar expressions. By incorporating the $\mathit{CICScores}$ and $\mathit{AblScores}$ terminals into G2, \sycam can generate more diverse expressions.
\Cref{fig:sanitycheck} illustrates the results obtained by \sycam using both G1 and G2. When G1 was used, \sycam managed to synthesize GradCAM++, represented by $\mathit{ReLU(Grads)}$ \cite{lerma2022gradcam}, guided by the Average Drop \% metric, and did not synthesize a better expression before the timeout. However, when G2 was used, \sycam was able to synthesize even better expressions, with a reduction of the Average Drop \% metric of 5\%. This shows that \sycam can synthesize the dominant known expression, 
 or even better expressions if there is a possibility of doing so, and the grammar allows it.

\subsection{Experiments E2: Enumerate \& Class-based decomposition}
\label{section:enumvsclass}
\textit{"Show the efficacy of \sycam, and that the class-based decomposition approach can perform better than the enumerate approach."}

We use this set of experiments to show the efficacy of \sycam in comparison to other well-known CAM methods: GradCAM, GradCAM++, ScoreCAM, and AblationCAM. We also show how the class-based decomposition allows us to obtain better results than the simple enumerative approach. 

\begin{figure}[t]
    \centering
    \scalebox{0.9}{
\includegraphics[width=\linewidth]{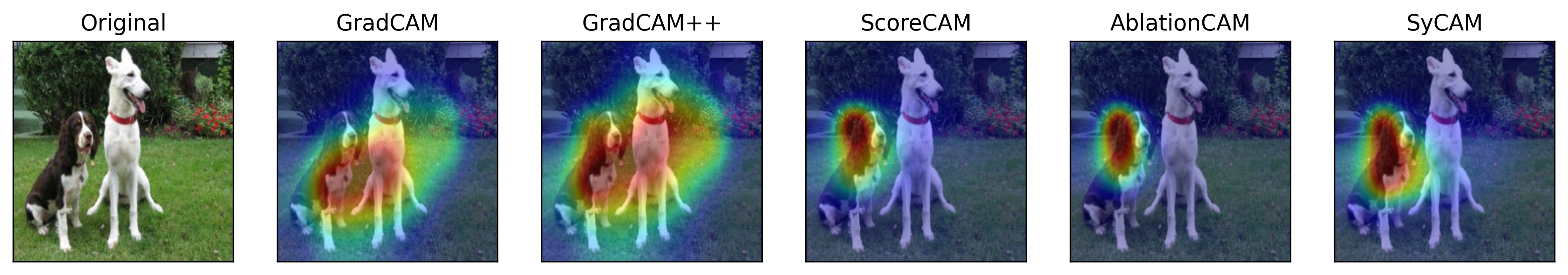}
    }
    \caption{
    \small Saliency maps generated by GradCAM, GradCAM++, ScoreCAM, AblationCAM, and the \sycam expression for ResNet50, the class "2. English springer", and the Deletion metric ($P=30$, higher is better). The scores for each method are 0.2141, 0.2414, 0.2419, 0.2343, and 0.2441, respectively. \sycam gets a better score and a saliency map that does not highlight the right dog as much as GradCAM and GradCAM++ do and highlights the body of the left dog more than ScoreCAM and AblationCAM.}
    \label{fig:heatmapdogs}
\end{figure}

In general, whilst the average value-wise improvement in the score obtained by \sycam in each experiment may look only marginally better than the base methods, this marginally better score already results in more targeted saliency maps for a significant number of images. In \Cref{fig:heatmapdogs}, while the \sycam score shows minor improvement, \sycam gets a better saliency map that does not highlight the right dog as GradCAM and GradCAM++ do and highlights the body of the left dog more than ScoreCAM and AblationCAM. This implies that the model is making the classification of the image mostly based on the left dog, specifically the head and the upper body. 

In the following, we give more details about our findings for both the enumerative and class-based decomposition approaches.

\paragraph{Results for the enumerative approach}  We use a reduced version of \textit{ImageNet}, namely the \textit{Imagenette} dataset \cite{imagette} that includes images for 10 classes out of the 1000 classes of \textit{ImageNet}. 
Specifically, we use the validation dataset, with 3925 images distributed evenly into the 10 output classes. The evaluation metric in all the experiments is the Deletion or the Insertion metric. The experiments run for a fixed time of 24 hours. Then, they return the best expression found. For the correctness oracle, we use three subsets of images, $I \supset I_2 \supset I_1$. The set ${I}_2$ is a subset of 1000 images: 100 from each class. ${I}_1$ contains 100 images: 10 per class. 

As explained in \Cref{section:experimentalsetup}, the evaluation of each candidate over the whole dataset can lead to scalability issues. Around 1020 candidate expressions are generated in three iterations of \Cref{alg:bottomupsearch}, and it takes ${\sim}0.3$s to evaluate the Deletion or the Insertion metric ($P=10$) over a single image. The computation time of the evaluation of a candidate expression over each subset for the Deletion and Insertion metrics ($P=10$) is ${\sim}18$m, ${\sim}5$m, and ${\sim}27$s, respectively.

\begin{table}[!t]
\centering
\caption{\small Enumerative approach for the models ResNet50, VGG-16, and VGG-19 over the Imagenette dataset using our variants of the Deletion and  Insertion metrics (higher is better). G=GradCAM, G+=GradCAM++, S=ScoreCAM, A=AblationCAM.}
\label{Tab:enumerative}
\makebox[\linewidth][c]{
\scalebox{0.72}{
\begin{tabular}{ c c c c c c c c c c c c c }
    \toprule 
    \multicolumn{1}{c}{} & \multicolumn{6}{c}{\small Deletion metric (P=10)} & \multicolumn{6}{c}{\small Insertion metric (P=10)}   \\ \cmidrule(l){2-7} \cmidrule(l){8-13}
    Model & G & G+  & S & A & \sycam & \sycam Exp. &  G & G+  & S & A & \sycam & \sycam Exp.  \\
    \hline
    ResNet50 & 0.3559 & 0.3479 & 0.3523 & 0.3601 & \textbf{0.3619} & \begin{tabular}{c}$5 \cdot \mathit{Grads}$ \\ $+ \ \mathit{top}_{50}$ \\ $+ \ \mathit{ReLU(Grads)}$ \\ $+ \ \mathit{AblScores}$\end{tabular} & 0.2287 & 0.2204 & 0.2243 & 0.2219 & \textbf{0.2296} & \begin{tabular}{c}$\mathit{top}_{50}$ \\ $* \ \mathit{CICScores}$\end{tabular} \\ 
    \hline
    VGG-16 & 0.1722 & 0.1718 & 0.1727 & 0.1836 &  \textbf{0.1883} & \begin{tabular}{c}$2 \cdot \mathit{Grads}$ \\ $+ \  \mathit{top}_5$ \\ $+ \  \mathit{CICScores}$ \\ $+ \  \mathit{AblScores}$\end{tabular} & 0.0167 & 0.0185 & \textbf{0.0320} & 0.0204 & \textbf{0.0320} & $\mathit{CICScores}$ \\
    \hline
    VGG-19 & 0.1736 & 0.1713 & 0.1744 & 0.1824 & \textbf{0.1825} & \begin{tabular}{c}$\mathit{top}_{10}$ \\ $+ \ \mathit{AblScores}$\end{tabular} & 0.0260 & 0.0210 & \textbf{0.0346} & 0.0237 & \textbf{0.0346} & $\mathit{CICScores}$ \\ \bottomrule 
\end{tabular}
}} 
\label{Tab:enumerate_L10}

\makebox[\linewidth][c]{
\scalebox{0.71}{
\begin{tabular}{ c c c c c c c c c c c c c }
    \toprule 
    \multicolumn{1}{c}{} & \multicolumn{6}{c}{Deletion metric (P=30)} & \multicolumn{6}{c}{Insertion metric (P=30)}   \\ \cmidrule(l){2-7} \cmidrule(l){8-13}
    Model & G & G+  & S & A & \sycam & \sycam Exp. &  G & G+  & S & A & \sycam & \sycam Exp.  \\
    \hline 
    ResNet50 & 0.6075 & 0.6010 & 0.6017 & 0.6049 & \textbf{0.6079} & \begin{tabular}{c}$\mathit{Grads}$ \\ $+ \mathit{top}_{10}$\end{tabular} & \textbf{0.5435} & 0.5289 & 0.5349 & 0.5402 & \textbf{0.5435} & $\mathit{Grads}$ \\ 
    \hline
    VGG-16 & 0.3344 & 0.3382 & 0.3385 & 0.3523 & \textbf{0.3579} & \begin{tabular}{c}$4 \cdot \mathit{Grads}$ \\ $+ \ \mathit{top}_{5}$ \\ $+ \ \mathit{CICScores}$ \\ $+ \ \mathit{AblScores}$\end{tabular} & 0.0957 & 0.1075 & \textbf{0.1422} & 0.1160 & \textbf{0.1422} & $\mathit{CICScores}$ \\
    \hline
    VGG-19 & 0.3337 & 0.3364 & 0.3340 & \textbf{0.3454} & \textbf{0.3454} & $\mathit{AblScores}$ & 0.0996 & 0.1105 & 0.1422 & 0.1146 & \textbf{0.1424} & \begin{tabular}{c}$2 \cdot \mathit{Grads}$ \\ $+ \ \mathit{ReLU(Grads)}$ \\ $+ \ \mathit{CICScores}$ \end{tabular} \\ \bottomrule 
\end{tabular}
}}
\label{Tab:enumerate_L30}
\end{table}

\begin{table}[!htb]
\centering
\caption{\small Class-based decomposition approach for ResNet50 over the Imagenette dataset using the Deletion and Insertion metrics (higher is better). G=GradCAM, G+=GradCAM++, S=ScoreCAM, A=AblationCAM.}
\label{Tab:resnet50_divide}
\makebox[\linewidth][c]{
\scalebox{0.67}{
\begin{tabular}{ c c c c c c c c c c c c c }
    \toprule 
    \multicolumn{1}{c}{} & \multicolumn{6}{c}{Deletion metric (P=10)} & \multicolumn{6}{c}{Insertion metric (P=10)}   \\ \cmidrule(l){2-7} \cmidrule(l){8-13}
    Model & G & G+  & S & A & \sycam & \sycam Exp. &  G & G+  & S & A & \sycam & \sycam Exp.  \\
    \hline  
    1.Tench & 0.4329 & 0.4286 & 0.4203 & 0.4258 & \textbf{0.4358} & \begin{tabular}{c}$\mathit{Grads}$ \\ $+ \mathit{top}_{5}$\end{tabular} & 0.2307 & 0.2209 & 0.2188 & 0.2346 & \textbf{0.2367} & \begin{tabular}{c}$\mathit{CICScores}$ \\ $* \mathit{AblScores}$\end{tabular}\\ 
    \hline
    2.English springer & 0.4293 & 0.4193 & 0.4413 & \textbf{0.4513} &  0.4406 & \begin{tabular}{c}$\mathit{Grads}^2$ \\ $+ \mathit{top}_{50}$ \\ $* \mathit{CICScores}$ \end{tabular} & 0.1741 & 0.1639 & \textbf{0.1786} & 0.1706 & 0.1741 & $\mathit{Grads}$ \\
    \hline
    3.Cassette player & 0.2755 & 0.2719 & 0.2698 & 0.2577 & \textbf{0.2766} & $\mathit{top}_{50}$ & 0.1584 & 0.1517 & 0.1514 & 0.1201 & \textbf{0.1663} & $\mathit{top}_{10}$ \\ 
    \hline
    4.Chain saw & 0.4360 & 0.4301 & \textbf{0.4439} & 0.4432 & 0.4404 & $\mathit{top}_{10}$ & 0.2524 & 0.2425 & \textbf{0.2536} & 0.2327 & 0.2524 & $\mathit{Grads}$ \\ 
    \hline
    5.Church & 0.2634 & 0.2468 & 0.2505 & \textbf{0.2799} & 0.2640 & \begin{tabular}{c}$2 \cdot \mathit{Grads}$ \\ $+ \mathit{top}_{20}$\end{tabular} & 0.0859 & 0.0803 & 0.0880 & 0.0862 & \textbf{0.089}1 & $\mathit{top}_{20}$ \\ 
    \hline
    6.French horn & 0.3891 & 0.3849 & 0.3831 & \textbf{0.3972} & 0.3891 & $\mathit{Grads}$ & 0.2550 & 0.2441 & 0.2460 & \textbf{0.2693} & 0.2550 & $\mathit{Grads}$ \\ 
    \hline
    7.Garbage truck & 0.3757 & 0.3643 & 0.3650 & \textbf{0.3779} &  0.3757 & $\mathit{Grads}$ & 0.2411 & 0.2390 & 0.2412 & 0.2328 & \textbf{0.2438} & \begin{tabular}{c}$\mathit{Grads}$ \\ $+ \mathit{top}_{20}$ \\ $+ \mathit{Grads}^2$\end{tabular} \\ 
    \hline
    8.Gas Pump & 0.3405 & 0.3262 & 0.3369 & \textbf{0.3470} & 0.3405 & $\mathit{Grads}$ & 0.0508 & 0.0472 & \textbf{0.0570} & 0.0464 & 0.0508 & $\mathit{Grads}$ \\
    \hline
    9.Golf ball & 0.3309 & 0.3272 & 0.3206 & 0.3279 & \textbf{0.3356} & $\mathit{top}_{10}$ & 0.4149 & 0.4047 & 0.3953 & \textbf{0.4200} & 0.4167 & $\mathit{top}_{50}$ \\ 
    \hline
    10.Parachute & 0.2855 & 0.2805 & \textbf{0.2918} & 0.2897 & 0.2858 &  \begin{tabular}{c}$2 \cdot \mathit{Grads}$ \\ $+ \mathit{top}_{10}$\end{tabular} & \textbf{0.4343} & 0.4205 & 0.4231 & 0.4129 & \textbf{0.4343} & $\mathit{Grads}$ \\  \hline
    Average & 0.3559 & 0.3480 & 0.3523 & \textbf{0.3598} & 0.3584 & & 0.2287 & 0.2215 & 0.2253 & 0.2226 & \textbf{0.2308} & \\ \bottomrule 
\end{tabular}
}}
\label{Tab:resnet_L10}

\makebox[\linewidth][c]{
\scalebox{0.68}{
\begin{tabular}{ c c c c c c c c c c c c c }
    \toprule 
    \multicolumn{1}{c}{} & \multicolumn{6}{c}{ Deletion metric (P=30)} & \multicolumn{6}{c}{Insertion metric (P=30)}   \\ \cmidrule(l){2-7} \cmidrule(l){8-13}
    Model & G & G+  & S & A & \sycam & \sycam Exp. &  G & G+  & S & A & \sycam & \sycam Exp.  \\
    \hline 
    1.Tench & 0.7215 & 0.7170 & 0.7120 & 0.7148 & \textbf{0.7236} & \begin{tabular}{c}$ \mathit{top}_{50} $ \\ $ * \mathit{CICScores}$\end{tabular} & 0.6415 & 0.6310 & 0.6297 & 0.6398 & \textbf{0.6415} & $\mathit{Grads}$ \\ 
    \hline
    2.English springer & 0.6360 & 0.6290 & 0.6365 & \textbf{0.6447} &  0.6416 & \begin{tabular}{c}$ \mathit{top}_{50} $ \\ $ * \mathit{CICScores}$\end{tabular} & 0.4815 & 0.4672 & 0.4746 & \textbf{0.4855} & 0.4815 & $\mathit{Grads}$ \\
    \hline
    3.Cassette player & 0.4374 & 0.4336 & 0.4323 & 0.4169 & \textbf{0.4385} & $\mathit{top}_{10}$ & 0.3404 & 0.3256 & 0.3324 & 0.2932 & \textbf{0.3427} & \begin{tabular}{c}$\mathit{Grads}$ \\ $+ \mathit{top}_{5}$\end{tabular} \\ 
    \hline
    4.Chain saw & 0.6711 & 0.6667 & 0.6694 & 0.6504 & \textbf{0.6730} & $\mathit{top}_{10}$ & 0.5896 & 0.5698 & 0.5863 & 0.5678 & \textbf{0.5896} & $\mathit{Grads}$ \\ 
    \hline
    5.Church & 0.4677 & 0.4548 & 0.4580 & \textbf{0.4755} & 0.4692 & \begin{tabular}{c}$2 \cdot \mathit{Grads}$ \\ $+ \mathit{top}_{20}$\end{tabular} & 0.3488 & 0.3319 & 0.3442 & \textbf{0.3632} & 0.3501 & \begin{tabular}{c}$2 \cdot \mathit{Grads}$ \\ $+ \mathit{top}_{20}$\end{tabular} \\ 
    \hline
    6.French horn & 0.6464 & 0.6412 & 0.6363 & \textbf{0.6527} & 0.6476 & $\mathit{top}_{50}$ & 0.5968 & 0.5816 & 0.5867 & \textbf{0.6126} & 0.5968 & $\mathit{Grads}$ \\
    \hline
    7.Garbage truck & 0.6676 & 0.6619 & 0.6622 & \textbf{0.6681} &  0.6676 & $\mathit{Grads}$ & 0.6072 & 0.5972 & 0.6030 & \textbf{0.6091} & 0.6072 & $\mathit{Grads}$ \\ 
    \hline
    8.Gas Pump & 0.6286 & 0.6223 & 0.6288 & \textbf{0.6341} & 0.6286 & $\mathit{Grads}$ & 0.4049 & 0.3862 & 0.4008 & 0.4045 & \textbf{0.4050} & \begin{tabular}{c}$2 \cdot \mathit{Grads}$ \\ $+ \mathit{top}_{5}$\end{tabular} \\
    \hline
    9.Golf ball & 0.5865 & 0.5791 & 0.5649 & 0.5875 & \textbf{0.5895} & $\mathit{top}_{10}$ & 0.7158 & 0.7072 & 0.6987 & \textbf{0.7224} & 0.7158 & $\mathit{Grads}$ \\ 
    \hline
    10.Parachute & 0.6041 & 0.5972 & \textbf{0.6090} & 0.5930 & 0.6041 &  $\mathit{Grads} $ & 0.7084 & 0.6924 & 0.6928 & 0.6982 & \textbf{0.7084} & $\mathit{Grads}$ \\  \hline 
    Average & 0.6075 & 0.6010 & 0.6017 & 0.6049 & \textbf{0.6091} & & 0.5435 & 0.5289 & 0.5349 & 0.5402 & \textbf{0.5438} &   \\ \bottomrule 
\end{tabular}
}}
\label{Tab:resnet_L30}
\end{table}

In \Cref{Tab:enumerative}, we present our results for all three models. We show the expressions synthesized by \sycam, and provide a comparison with GradCAM, GradCAM++, ScoreCAM, and AblationCAM,
showing the average scores for each method when evaluated using the Deletion and the Insertion metrics.
In general, \sycam synthesizes expressions better than the base methods. In some cases, \sycam synthesizes GradCAM ($\mathit{Grads}$), ScoreCAM ($\mathit{CICScores}$), or AblationCAM ($\mathit{AblScores}$). We point to the fact that AblationCAM performs better than the rest of the base methods for the Deletion metric, but not for the Insertion metric, for which it is almost always surpassed by ScoreCAM. This emphasize the necessity of using a framework like \sycam to synthesize the best expression for each context.
For $P=15$ and $P=30$, the computation time grows proportionally to $P$. There are some exceptions because of the correctness oracle. In the worst case, every expression is better than the threshold for $I_1$ and $I_2$ but worse than $I$ and must be evaluated over the three sets. 
Note that both metrics depend on the ratio between $P$ and the size of the feature maps in the last convolutional layer. Because the feature maps are bigger in VGG-16 and VGG-19 (14x14) than in ResNet50 (7x7), this dependency makes the values obtained by both metrics larger for ResNet50 than for VGG-19 and VGG-16 when comparing the same CAM methods. Larger values of $P$ are correlated with better metric scores when comparing the same CAM methods.


\paragraph{Results for the Class-based decomposition approach} 
For each of the ten classes, the correctness oracle will use a subset $I$, which contains approximately 390 images each, and ${I}_1 \subset I$, which contains 100 images each. The time needed for evaluating an expression over each subset is ${\sim}2$m$10$s and ${\sim} 30$s, respectively. 

To make a fair comparison between the enumerate and the class-based decomposition approach, we established a timeout of 2.4 hours per class. Although this limits the size of the expressions synthesized for each class, we can see in \Cref{Tab:resnet50_divide} that, for ResNet50, the average scores of \sycam per class show improvement over the enumerative approach except for the Deletion metric for L=10 case. 
For certain images, \sycam resulted in scores higher than those of GradCAM, GradCAM++, ScoreCAM, and AblationCAM. In other cases, \sycam was outperformed. This is because other methods may get much higher scores for a small subset of the data, but not for the rest. As explained in \Cref{section:experimentalsetup}, our goal is to find expressions that are better than the best one found in at least half of the images and that have a higher average score. 

\subsection{Experiments E3: Incorporating ground truth}
\textit{"Show how to incorporate expert knowledge into the generation of saliency maps."}

    We use the COVID-QU-Ex dataset \cite{covid_dataset}, which consists of 5826 chest X-ray images with infection segmentation data distributed between 1456 normal (healthy) images, 2913 COVID-19 images, and 1457 non-COVID images with other diseases. 
    Ground-truth COVID-19 segmentation masks are provided. We fine-tuned a ResNet50 model 
    to correctly classify the test dataset with accuracy higher than 95\%, and precision and recall higher than 92\% for each class. 
    
    We consider a set $I$ with only the 2913 COVID-19 X-ray images and the associated ground-truth COVID-19 segmentation masks. For the correctness oracle, we use three subsets of images, $I \supset I_2 \supset I_1$. The set ${I}_2$ is a subset of 1000 images, while ${I}_1$ contains 100 images. We apply \sycam using two metrics: the $m_{GT}$ and the SCH metrics defined in \Cref{subsection:evaluationmetric}. Because these metrics give importance to the ground truth COVID-19 masks, the \sycam expressions incorporate expert knowledge into the saliency maps generated. 
    Since we are only considering a binary classification, we employ the enumerative approach. For both metrics defined above, we let the experiments run for 24 hours. In this experiment, evaluating $m_{GT}$ or SCH takes around $0.05$s per image. 
    
    The results are presented in \Cref{Tab:enumerate_xray}. We can observe that, for both metrics, \sycam synthesizes expressions whose basis is of the form $exp^n$, with $exp$ a terminal and $n \in \mathbb{N}$: $\mathit{CICScores}^4$ for the $m_{GT}$ metric, and $\mathit{Grads}^5$ for the SCH metric. This may indicate that dominant expressions have this form, which highly reduces the values lower than 1 and gives more importance to the high values.
    
\begin{figure}
    \centering
    \begin{subfigure}[t]{0.8\textwidth}
        \centering
        \includegraphics[width=\linewidth]{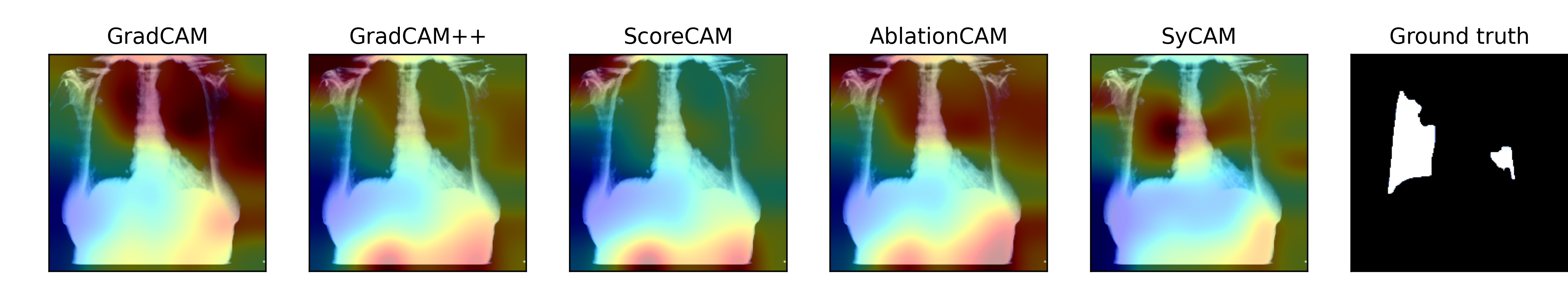}
        \caption{
        \small Heatmaps generated for the $m_{GT}$ metric (higher is better). The score for each method is 0.0 for every method except for \sycam, for which it is 0.3389. The \sycam heatmap is more concentrated on the ground truth mask.}
        \label{fig:heatmapxraymgt}
    \end{subfigure}
    
    \begin{subfigure}[t]{0.8\textwidth}
        \centering
        \includegraphics[width=\linewidth]{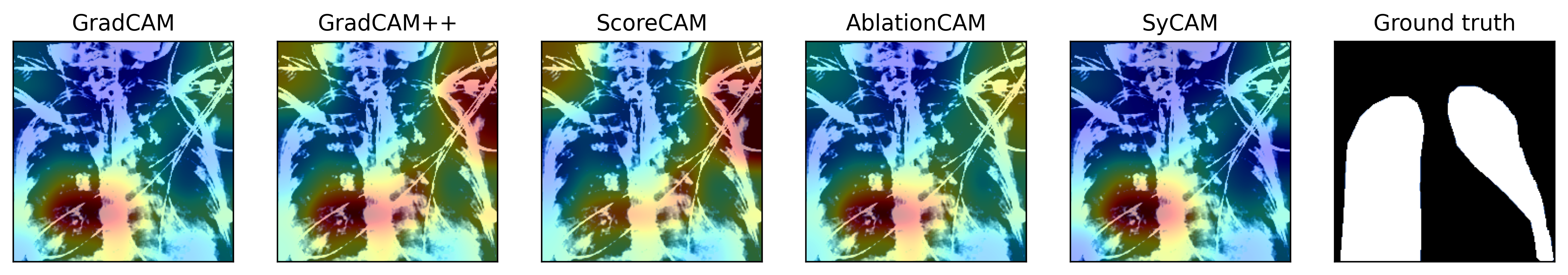}
        \caption{
        \small Heatmaps generated for the SCH metric (higher is better). The scores for each method are 0.3474, 0.2874, 0.2650, 0.3377, and 0.3803, respectively. The \sycam heatmap is a more compact version of GradCAM.}
        \label{fig:heatmapxraysch}
    \end{subfigure}
    \caption{\small
    Saliency maps generated by GradCAM, GradCAM++, ScoreCAM, AblationCAM, and the \sycam expression synthesized for ResNet50.
    } 
\end{figure}

    Examples of saliency maps generated by \sycam are included in \Cref{fig:heatmaps,fig:heatmapxraymgt,fig:heatmapxraysch}. We can observe in the X-ray images of \Cref{fig:heatmaps} that the saliency map generated by \sycam is more concentrated on the ground-truth mask, while the rest of the CAM methods fail to do so. In fact, we can see that GradCAM++ and ScoreCAM give high importance to a region outside of the body in the bottom right. Something similar happens in \Cref{fig:heatmapxraymgt}. In \Cref{fig:heatmapxraysch}, because the \sycam expression is $2 \cdot \mathit{Grads}^5$, the heatmap obtained is a more compact version of the GradCAM heatmap. This reduces the importance of areas outside of the ground truth, achieving a higher SCH score.  

\begin{table}[!t]
\centering
\caption{\small Enumerative approach for the fine-tuned model ResNet50 over the COVID-19 benchmark. The metrics used are the $m_{GT}$ and the SCH metrics (higher is better). }
\label{Tab:enumerate_xray}
\makebox[\linewidth][c]{
\scalebox{0.8}{
\begin{tabular}{ c c c c c c c }
    \toprule 
    Metric & GradCAM & GradCAM++ & ScoreCAM & AblationCAM & \sycam & \sycam Exp.  \\
    \hline 
    $m_{GT}$ & 0.1861 & 0.1712 & 0.1512 & 0.1876 & \textbf{0.1972} & \begin{tabular}{c}$\mathit{Grads}$ \\ $* \mathit{top}_{20}$ \\ $* CIC^4$\end{tabular} \\
    \hline 
    SCH & 0.1567 & 0.1491 & 0.1458 & 0.1524 & \textbf{0.1610} & $2 \cdot \mathit{Grads}^5$   \\ \bottomrule 
    
\end{tabular}
}}
\end{table} 

\section{Discussion and Future Work}


We presented \sycam, a metric-based synthesis framework for automatically generating CAM expressions. \sycam offers advantages in tailoring CAM expressions to specific syntactic restrictions, datasets, and evaluation metrics. However, our approach still has limitations that we plan to address in future work.
One key limitation is the high computation time required to generate expressions, which stems from the complexity of the synthesis algorithms. Improvements in synthesis methods will directly enhance our approach. However, we emphasize that generating a \sycam expression is a one-time process, and the computation time for generating saliency maps remains reasonable.
Additionally, the \sycam framework relies on enumerating expressions and verifying their validity. Future work will explore ways to guide this search process more efficiently, thereby reducing computation time.
Lastly, we also aim to extend our study to multi-objective settings \cite{DBLP:conf/fmcad/TorfahSCAS21}, incorporating multiple evaluation metrics simultaneously and identifying Pareto-optimal expressions as solutions.


\bibliographystyle{splncs04}
\bibliography{ref.bib}

\begin{thebibliography}{10}
\providecommand{\url}[1]{\texttt{#1}}
\providecommand{\urlprefix}{URL }
\providecommand{\doi}[1]{https://doi.org/#1}

\bibitem{bottomupsearch}
Albarghouthi, A., Gulwani, S., Kincaid, Z.: Recursive program synthesis. In:
  Sharygina, N., Veith, H. (eds.) Computer Aided Verification - 25th
  International Conference, {CAV} 2013, Saint Petersburg, Russia, July 13-19,
  2013. Proceedings. Lecture Notes in Computer Science, vol.~8044, pp.
  934--950. Springer (2013). \doi{10.1007/978-3-642-39799-8\_67}

\bibitem{sygus}
Alur, R., Bod{\'{\i}}k, R., Juniwal, G., Martin, M.M.K., Raghothaman, M.,
  Seshia, S.A., Singh, R., Solar{-}Lezama, A., Torlak, E., Udupa, A.:
  Syntax-guided synthesis. In: Formal Methods in Computer-Aided Design, {FMCAD}
  2013, Portland, OR, USA, October 20-23, 2013. pp.~1--8. {IEEE} (2013),
  \url{https://ieeexplore.ieee.org/document/6679385/}

\bibitem{clevr}
Arras, L., Osman, A., Samek, W.: {CLEVR-XAI:} {A} benchmark dataset for the
  ground truth evaluation of neural network explanations. Inf. Fusion
  \textbf{81},  14--40 (2022). \doi{10.1016/J.INFFUS.2021.11.008}

\bibitem{GradCAM++}
Chattopadhyay, A., Sarkar, A., Howlader, P., Balasubramanian, V.N.:
  {Grad-CAM++}: Generalized gradient-based visual explanations for deep
  convolutional networks. In: 2018 {IEEE} Winter Conference on Applications of
  Computer Vision, {WACV} 2018, Lake Tahoe, NV, USA, March 12-15, 2018. pp.
  839--847. {IEEE} Computer Society (2018). \doi{10.1109/WACV.2018.00097}

\bibitem{imagenet}
Deng, J., Dong, W., Socher, R., Li, L., Li, K., Fei{-}Fei, L.: Imagenet: {A}
  large-scale hierarchical image database. In: 2009 {IEEE} Computer Society
  Conference on Computer Vision and Pattern Recognition {(CVPR} 2009), 20-25
  June 2009, Miami, Florida, {USA}. pp. 248--255. {IEEE} Computer Society
  (2009). \doi{10.1109/CVPR.2009.5206848}

\bibitem{ablationcam}
Desai, S., Ramaswamy, H.G.: {Ablation-CAM}: Visual explanations for deep
  convolutional network via gradient-free localization. In: {IEEE} Winter
  Conference on Applications of Computer Vision, {WACV} 2020, Snowmass Village,
  CO, USA, March 1-5, 2020. pp. 972--980. {IEEE} (2020).
  \doi{10.1109/WACV45572.2020.9093360}

\bibitem{cem}
Dhurandhar, A., Chen, P., Luss, R., Tu, C., Ting, P., Shanmugam, K., Das, P.:
  Explanations based on the missing: Towards contrastive explanations with
  pertinent negatives. In: Bengio, S., Wallach, H.M., Larochelle, H., Grauman,
  K., Cesa{-}Bianchi, N., Garnett, R. (eds.) Advances in Neural Information
  Processing Systems 31: Annual Conference on Neural Information Processing
  Systems 2018, NeurIPS 2018, December 3-8, 2018, Montr{\'{e}}al, Canada. pp.
  590--601 (2018),
  \url{https://proceedings.neurips.cc/paper/2018/hash/c5ff2543b53f4cc0ad3819a36752467b-Abstract.html}

\bibitem{pascal-voc-2007}
Everingham, M., Van~Gool, L., Williams, C.K.I., Winn, J., Zisserman, A.: The
  {PASCAL} {V}isual {O}bject {C}lasses {C}hallenge 2007 {(VOC2007)} {R}esults.
  http://www.pascal-network.org/challenges/VOC/voc2007/workshop/index.html

\bibitem{schmetric}
Fletcher, L., van~der Klis, R., Sedl{\'{a}}cek, M., Vasilev, S., Athanasiadis,
  C.: Reproducibility study of "{LICO}: Explainable models with language-image
  consistency"  (2024), \url{https://doi.org/10.48550/arXiv.2410.13989}

\bibitem{XGradCAM}
Fu, R., Hu, Q., Dong, X., Guo, Y., Gao, Y., Li, B.: Axiom-based {Grad-CAM}:
  Towards accurate visualization and explanation of {CNNs}. In: 31st British
  Machine Vision Conference 2020, {BMVC} 2020, Virtual Event, UK, September
  7-10, 2020. {BMVA} Press (2020),
  \url{https://www.bmvc2020-conference.com/assets/papers/0631.pdf}

\bibitem{6909475}
Girshick, R., Donahue, J., Darrell, T., Malik, J.: Rich feature hierarchies for
  accurate object detection and semantic segmentation. In: 2014 IEEE Conference
  on Computer Vision and Pattern Recognition (CVPR). pp. 580--587. IEEE
  Computer Society, Los Alamitos, CA, USA (jun 2014).
  \doi{10.1109/CVPR.2014.81}

\bibitem{artifacts}
Hooker, S., Erhan, D., Kindermans, P., Kim, B.: A benchmark for
  interpretability methods in deep neural networks. In: Wallach, H.M.,
  Larochelle, H., Beygelzimer, A., d'Alch{\'{e}}{-}Buc, F., Fox, E.B., Garnett,
  R. (eds.) Advances in Neural Information Processing Systems 32: Annual
  Conference on Neural Information Processing Systems 2019, NeurIPS 2019,
  December 8-14, 2019, Vancouver, BC, Canada. pp. 9734--9745 (2019),
  \url{https://proceedings.neurips.cc/paper/2019/hash/fe4b8556000d0f0cae99daa5c5c5a410-Abstract.html}

\bibitem{imagette}
Howard, J.: Imagenette, \url{https://github.com/fastai/imagenette/}

\bibitem{survey}
Ibrahim, R., Shafiq, M.O.: Explainable convolutional neural networks: {A}
  taxonomy, review, and future directions. {ACM} Comput. Surv.
  \textbf{55}(10),  206:1--206:37 (2023). \doi{10.1145/3563691}

\bibitem{ogis}
Jha, S., Seshia, S.A.: A theory of formal synthesis via inductive learning.
  Acta Informatica  \textbf{54}(7),  693--726 (2017).
  \doi{10.1007/S00236-017-0294-5}

\bibitem{XAItaxonomy}
Kadir, M.A., Mosavi, A., Sonntag, D.: Evaluation metrics for {XAI}: A review,
  taxonomy, and practical applications. In: 2023 IEEE 27th International
  Conference on Intelligent Engineering Systems (INES). pp. 000111--000124
  (2023). \doi{10.1109/INES59282.2023.10297629}

\bibitem{6909619}
Karpathy, A., Toderici, G., Shetty, S., Leung, T., Sukthankar, R., Fei-Fei, L.:
  Large-scale video classification with convolutional neural networks. In: 2014
  IEEE Conference on Computer Vision and Pattern Recognition. pp. 1725--1732
  (2014). \doi{10.1109/CVPR.2014.223}

\bibitem{medicalcnn}
Khalil, H., El-Hag, N., Sedik, A., El-Shafie, W., Mohamed, A.E.N., Khalaf,
  A.A.M., El-Banby, G.M., Abd El-Samie, F.I., El-Fishawy, A.S.: Classification
  of diabetic retinopathy types based on convolution neural network {(CNN)}.
  Menoufia Journal of Electronic Engineering Research
  \textbf{28}(ICEEM2019-Special Issue),  126--153 (2019).
  \doi{10.21608/mjeer.2019.76962}

\bibitem{554195}
Lawrence, S., Giles, C., Tsoi, A.C., Back, A.: Face recognition: a
  convolutional neural-network approach. IEEE Transactions on Neural Networks
  \textbf{8}(1),  98--113 (1997). \doi{10.1109/72.554195}

\bibitem{attentioncam}
Leem, S., Seo, H.: Attention guided {CAM:} visual explanations of vision
  transformer guided by self-attention. In: Wooldridge, M.J., Dy, J.G.,
  Natarajan, S. (eds.) Thirty-Eighth {AAAI} Conference on Artificial
  Intelligence, {AAAI} 2024, Thirty-Sixth Conference on Innovative Applications
  of Artificial Intelligence, {IAAI} 2024, Fourteenth Symposium on Educational
  Advances in Artificial Intelligence, {EAAI} 2014, February 20-27, 2024,
  Vancouver, Canada. pp. 2956--2964. {AAAI} Press (2024).
  \doi{10.1609/AAAI.V38I4.28077}

\bibitem{lerma2022gradcam}
Lerma, M., Lucas, M.: {Grad-CAM++} is equivalent to {Grad-CAM} with positive
  gradients  (2022). \doi{10.48550/ARXIV.2205.10838}

\bibitem{pytorch}
Marcel, S., Rodriguez, Y.: Torchvision the machine-vision package of torch. In:
  Bimbo, A.D., Chang, S., Smeulders, A.W.M. (eds.) Proceedings of the 18th
  International Conference on Multimedia 2010, Firenze, Italy, October 25-29,
  2010. pp. 1485--1488. {ACM} (2010). \doi{10.1145/1873951.1874254}

\bibitem{fidelity}
Miró-Nicolau, M., i~Capó, A.J., Moyà-Alcover, G.: Assessing fidelity in xai
  post-hoc techniques: A comparative study with ground truth explanations
  datasets. Artificial Intelligence  \textbf{335},  104179 (2024).
  \doi{https://doi.org/10.1016/j.artint.2024.104179}

\bibitem{augmentedGC}
Morbidelli, P., Carrera, D., Rossi, B., Fragneto, P., Boracchi, G.: Augmented
  {Grad-CAM}: Heat-maps super resolution through augmentation. In: 2020 {IEEE}
  International Conference on Acoustics, Speech and Signal Processing, {ICASSP}
  2020, Barcelona, Spain, May 4-8, 2020. pp. 4067--4071. {IEEE} (2020).
  \doi{10.1109/ICASSP40776.2020.9054416}

\bibitem{eigenCAM}
Muhammad, M.B., Yeasin, M.: {Eigen-CAM}: Visual explanations for deep
  convolutional neural networks. {SN} Comput. Sci.  \textbf{2}(1), ~47 (2021).
  \doi{10.1007/S42979-021-00449-3}

\bibitem{smoothgradpp}
Omeiza, D., Speakman, S., Cintas, C., Weldemariam, K.: Smooth {Grad-CAM++}: An
  enhanced inference level visualization technique for deep convolutional
  neural network models  (2019), \url{http://arxiv.org/abs/1908.01224}

\bibitem{rise}
Petsiuk, V., Das, A., Saenko, K.: {RISE:} randomized input sampling for
  explanation of black-box models. In: British Machine Vision Conference 2018,
  {BMVC} 2018, Newcastle, UK, September 3-6, 2018. p.~151. {BMVA} Press (2018),
  \url{http://bmvc2018.org/contents/papers/1064.pdf}

\bibitem{rafferty2024transparentclinicallyinterpretableai}
Rafferty, A., Ramaesh, R., Rajan, A.: Transparent and clinically interpretable
  {AI} for lung cancer detection in chest {X}-rays (2024),
  \url{https://arxiv.org/abs/2403.19444}

\bibitem{IROF}
Rieger, L., Hansen, L.K.: {IROF:} a low resource evaluation metric for
  explanation methods  (2020), \url{https://arxiv.org/abs/2003.08747}

\bibitem{road}
Rong, Y., Leemann, T., Borisov, V., Kasneci, G., Kasneci, E.: A consistent and
  efficient evaluation strategy for attribution methods. In: Chaudhuri, K.,
  Jegelka, S., Song, L., Szepesv{\'{a}}ri, C., Niu, G., Sabato, S. (eds.)
  International Conference on Machine Learning, {ICML} 2022, 17-23 July 2022,
  Baltimore, Maryland, {USA}. Proceedings of Machine Learning Research,
  vol.~162, pp. 18770--18795. {PMLR} (2022),
  \url{https://proceedings.mlr.press/v162/rong22a.html}

\bibitem{AOPC}
Samek, W., Binder, A., Montavon, G., Lapuschkin, S., M{\"{u}}ller, K.:
  Evaluating the visualization of what a deep neural network has learned.
  {IEEE} Trans. Neural Networks Learn. Syst.  \textbf{28}(11),  2660--2673
  (2017). \doi{10.1109/TNNLS.2016.2599820}

\bibitem{GradCAM}
Selvaraju, R.R., Cogswell, M., Das, A., Vedantam, R., Parikh, D., Batra, D.:
  Grad-cam: Visual explanations from deep networks via gradient-based
  localization. In: 2017 IEEE International Conference on Computer Vision
  (ICCV). pp. 618--626 (2017). \doi{10.1109/ICCV.2017.74}

\bibitem{Szczepankiewicz2023}
Szczepankiewicz, K., Popowicz, A., Charkiewicz, K.,
  Na{\l}{\k{e}}cz-Charkiewicz, K., Szczepankiewicz, M., Lasota, S.,
  Zawistowski, P., Radlak, K.: Ground truth based comparison of saliency maps
  algorithms. Scientific Reports  \textbf{13}(1),  16887 (Oct 2023).
  \doi{10.1038/s41598-023-42946-w}

\bibitem{covid_dataset}
Tahir, A.M., Chowdhury, M.E.H., Qiblawey, Y., Khandakar, A., Rahman, T.,
  Kiranyaz, S., Khurshid, U., Ibtehaz, N., Mahmud, S., Ezeddin, M.:
  Covid-qu-ex. Kaggle  (2021). \doi{10.34740/3122958}

\bibitem{DBLP:conf/fmcad/TorfahSCAS21}
Torfah, H., Shah, S., Chakraborty, S., Akshay, S., Seshia, S.A.: Synthesizing
  pareto-optimal interpretations for black-box models. In: Formal Methods in
  Computer Aided Design, {FMCAD} 2021, New Haven, CT, USA, October 19-22, 2021.
  {IEEE} (2021), \url{https://doi.org/10.34727/2021/isbn.978-3-85448-046-4\_24}

\bibitem{scoreCAM}
Wang, H., Wang, Z., Du, M., Yang, F., Zhang, Z., Ding, S., Mardziel, P., Hu,
  X.: {Score-CAM}: Score-weighted visual explanations for convolutional neural
  networks. In: 2020 Conference on Computer Vision and Pattern Recognition,
  {CVPR} Workshops 2020, Seattle, WA, USA, June 14-19, 2020. pp. 111--119.
  Computer Vision Foundation / {IEEE} (2020).
  \doi{10.1109/CVPRW50498.2020.00020}

\bibitem{CAM}
Zhou, B., Khosla, A., Lapedriza, {\`{A}}., Oliva, A., Torralba, A.: Learning
  deep features for discriminative localization. In: 2016 {IEEE} Conference on
  Computer Vision and Pattern Recognition, {CVPR} 2016, Las Vegas, NV, USA,
  June 27-30, 2016. pp. 2921--2929. {IEEE} Computer Society (2016).
  \doi{10.1109/CVPR.2016.319}

\end{thebibliography}


\end{document}